\documentclass[preprint,12pt]{elsarticle}

\usepackage{amssymb, xcolor, hyperref, bm, subcaption, caption, multirow, algpseudocode, algorithm, multirow, amsmath}

\newcommand{\mli}[1]{\mathit{#1}}

\journal{Knowledge-Based Systems}

\begin{document}

\begin{frontmatter}



\title{A novel approach to generate datasets with XAI ground truth to evaluate image models}

\author[1,2]{Miquel Mir\'{o}-Nicolau}
\ead{miquel.miro@uib.es}
\author[1,2]{Antoni Jaume-i-Cap\'{o}}
\ead{antoni.jaume@uib.es}
\author[1,2]{Gabriel Moy\`{a}-Alcover} \corref{cor1}
\ead{gabriel.moya@uib.es}

\cortext[cor1]{Corresponding author}
\address[1]{UGiVIA Research Group, University of the Balearic Islands, Dpt. of Mathematics and Computer Science, 07122 Palma (Spain)}
\address[2]{Laboratory for Artificial Intelligence Applications (LAIA@UIB), University of the Balearic Islands, Dpt. of Mathematics and Computer Science, 07122 Palma (Spain)}



            
\begin{abstract}
With the increased usage of artificial intelligence (AI), it is imperative to understand how these models work internally. These needs have led to the development of a new field called eXplainable artificial intelligence (XAI). This field consists of on a set of techniques that allows us to theoretically determine the cause of the AI decisions. One main issue of XAI is how to verify the works on this field, taking into consideration the lack of ground truth (GT). In this study, we propose a new method to generate datasets with GT. We conducted a set of experiments that compared our GT with real model explanations and obtained excellent results confirming that our proposed method is correct.
\end{abstract}


\end{frontmatter}


\section{Introduction}

Deep neural networks have become a de facto standard to solve complex task as computer vision~\cite{krizhevsky2012imagenet} or natural language processing~\cite{vaswani2017attention}, obtaining astonishing results in most fields. These good results were reached through the combination of larger data sets and increase model complexity. However, this complexity also provoked a set of caveats. One of them is the difficulty to understand the causes of their results. This opacity of the deep learning models is known as the black-box problem. This limitation is addressed by eXplainable artificial intelligence (XAI), which aims to open this models and avoid the black-box problem. As stated by Barredo Arrieta \emph{et al.}~\cite{barredo_arrieta_explainable_2020} “XAI proposes creating a suite of ML techniques that 1) produce more explainable models while maintaining a high level of learning performance (e.g., prediction accuracy), and 2) enable humans to understand, appropriately trust, and effectively manage the emerging generation of artificially intelligent partners.”

XAI techniques have been widely used in multiple fields, as can be seen in the work of Murdoch \emph{et al.}~\cite{murdoch_definitions_2019}. In particular, and due to the ethical and regulatory concerns, XAI had been largely used in sensitive contexts as medicine, as can be seen in the recent published reviews of the state-of-art (\cite{miro2022evaluating}, \cite{VanderVelden2021}). These studies assess that most XAI related work is not validated. In this same direction, Miller~\cite{miller2019explanation} assessed that \textit{most of the work about explainability relies on the authors’ intuition, and an essential point is to have metrics that describe the overall explainability with the aim to compare different models regarding their level of explainability}. This problem is not only for the different XAI methods, as discussed by Adebayo \emph{et al.}~\cite{adebayo_sanity_2018}, but also for adjacent element of the field as fidelity metrics~\cite{tomsett2020sanity}. All these limitations arise from the fact that, in a real scenario, it is not possible to have a ground truth for the explanations, because, a priori, multiple explanation can be correct to obtain good results, however, only one of the possible explanations is the one used by the backbone AI model. 

Recently, a set of studies emerged that proposed to generate artificial datasets with ground truth for the explanations (\cite{CORTEZ20131}, \cite{arras2020ground}, \cite{guidotti_evaluating_2021}, and \cite{mamalakis2022investigating}). These datasets allowed to validate the different XAI techniques, without adding new sources of problems or having to take unproven assumptions. Cortez and Embrechts~\cite{CORTEZ20131} proposed a new XAI method and generated a synthetic dataset with a ground truth for the explanation as a benchmark. The dataset consisted of 1000 different samples of four random values. This dataset also contained a synthetic label that was built using a weighted sum of the four values. An AI model was trained to regress this synthetic label from the original data, and for this reason a well-trained model, weights each feature of the dataset with the weights of the sum that generates the label. These weights represented the explanation ground truth. Nonetheless, their proposal had two limitations: it was only applied to regression problems and only used with tabular data. Arras \emph{et al.}~\cite{arras2020ground} proposed a methodology for generating a synthetic attribution dataset of images. The proposed methodology was based on visual question answering (VQA) models. These models are specific types of AI models that can answer questions regarding the content of an image. Arras \emph{et al.}~\cite{arras2020ground} proposed asking questions depending only on one object from the image. They considered an explanation correct if only  the object of interest in the question was highlighted. Following this methodology, they compared 11 different XAI methods. Arras \emph{et al.}~\cite{arras2020ground} proposal was capable to handle images data. Nevertheless, only in a very specific problem, the VQA models. They also did not take into account the importance value of the explanations; instead, they only considered the position of the most important elements. Guidotti~\cite{guidotti_evaluating_2021} acknowledge the black-box models trained on datasets with GT for explanation become transparent models due to the availability of the real explanation. He proposed to establish a set of generators specifically designed for this type of dataset. In the case of images, these resulting datasets can be effectively utilized for classification tasks. However, it is important to note that this approach focuses solely on a single pattern. As a result, the evaluation of explanations is limited to their spatial location. Mamalakis \emph{et al.}~\cite{mamalakis2022investigating} named this type of datasets \textit{synthetic attribution benchmark} and formalized them. They defined them as the conjunction of three elements: \textit{synthetic input $X$,  function $F$ and synthetic output $Y$, the latter is the result of applying $F$ to $X$}. In addition, the methodology they proposed was used for classification problems, similarly to Guidotti~\cite{guidotti_evaluating_2021} proposal, but with multiple pattern for an explanation. The $X$ of this dataset is a set of binary images of circles and squares. $Y$ was one of two classes: the first one indicated more area of circles than squares, and the second indicated the contrary. With this information, they compared 12 methods. All 12 methods indicate the positive or negative of each pixel in a given class. Mamalakis \emph{et al.} \cite{mamalakis2022investigating} proposal can be used with multiple pattern classes, but it only verifies whether the features have a negative or positive impact on the classification. Similar to Arras \emph{et al.}~\cite{arras2020ground} and Guidotti~\cite{guidotti_evaluating_2021} did not consider the value of the explanation.

In this paper, we proposed a new methodology based on the previous works to build synthetic attribution benchmark datasets with a defined GT for image tasks. Our proposal overcome the limitations of the other proposals on the existing state-of-art: not taking into account the value of the explanation and being unable to use it for most typical image-related tasks. With our methodology, we use the exact explanation value as a ground truth, and we can use it indistinctly in regression and classification contexts. This dataset going to allow the community to verify the different XAI techniques used and future developments.

The rest of this paper is organized as follows. In the next section, we describe our methodology for building synthetic attributions benchmarks dataset. In Section~\ref{sec:experimental_setup}, we specify the experimental environment, and we describe the datasets, measures, metrics, and predictive models used for experimentation. Section~\ref{sec:results} discuss the results and comparison experiments obtained after applying the two datasets generated using the proposed method to multiple common XAI methods. Finally, in Section~\ref{sec:conclusion} we present the conclusions of the study.

\section{Method}

We based our proposal on the work of Cortez and Embrechts~\cite{CORTEZ20131}. As explained in the previous section, these authors generated a synthetic attribution benchmark (SAB) dataset for tabular data. The proposed methodology enhances their method and applies it to images. Similar to the original work, we used it for regression, but also for classification tasks. This was accomplished by changing how the label was generated. Instead of tabular features, we proposed to use numerical information found on the images, such as the number of times a particular pattern was found on an image.

\begin{figure}[!htb]
\begin{center}
	\includegraphics[width = 0.85\linewidth]{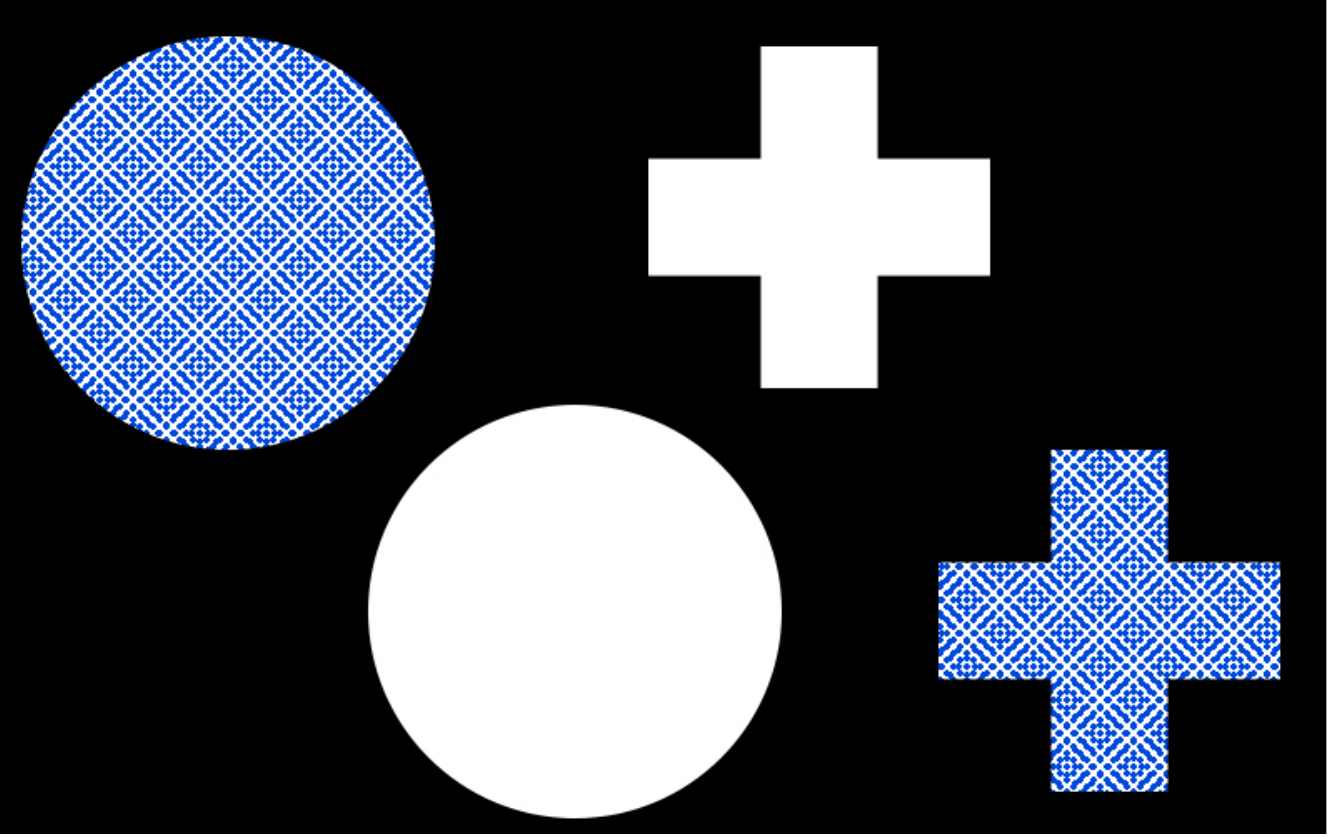}
\end{center}
\caption{Image containing different patterns. The total amount of patterns presented in the image is $|C| = 4$. Nonetheless, depending on the problem, only a subset of this $C$ is considered, e.g., if the problem is based on texture analysis, we have two different $p$, the uniform elements and the ones containing the blue texture.}\label{fig:patterns}
\end{figure}

To define our methodology, we use the formalization of SAB proposed by Mamalakis \emph{et al.}. We defined three different elements: the input data $\mathcal{X}$, attribution function $F$, and output data $\mathcal{Y}$. This was conducted as follows:

\noindent{\bf Input data:} In this step, we defined the input data, which is a collection of $n$ images: $\mathcal{X}= \{I_{1}, I_{2}, ... I_{n}\}$. In each image, we can find multiple occurrences of different patterns $p \in \mathcal{C}$, where $\mathcal{C}$ is the set of all possible patterns that can be identified in an image. The definition of $p$ is strongly related to the nature of the experiment. In figure \ref{fig:patterns}, for example, we have six different $p$, $C=\{$ circles, crosses, uniform figures, blue texture figures, uniform circles, uniform crosses, blue texture circles, and blue texture crosses $\}$. However, depending on the problem definition only a subset of $C$ can be considered.

\noindent{\bf Attribution function}. We define a  function $F:\mathcal{X} \to \mathbb{R}$. This function takes an image $I$ and uses the $p$ present on it, $F(I) = \sum_{i= 1}^{m} w_{i} \cdot g(p_{i}, I)$, where $m$ is the total number of different $p$ considered, $g(p_{i}, I)$ is a function that extracts some information of the pattern $p_{i}$ of the image $I$, for example the number of times that this pattern appears on the image, and $w_{i}$ is the ground truth for the importance of pattern $p_{i}$ for image $I$.

The attribution function can be either a continuous or a discrete function. The selection of which kind of function used depends on which is the task to evaluate (regression or classification). 

\noindent{\bf Output:}. We generated labels, or output data, by combining the function and input data. When $\mathcal{X}$ and $F$ are defined, the output data is $\mathcal{Y} = \{y: \forall I \in \mathcal{X}, y \Rightarrow F(I)\}$. 

\begin{figure}[!htb]
\begin{center}
	\includegraphics[width = 0.85\linewidth]{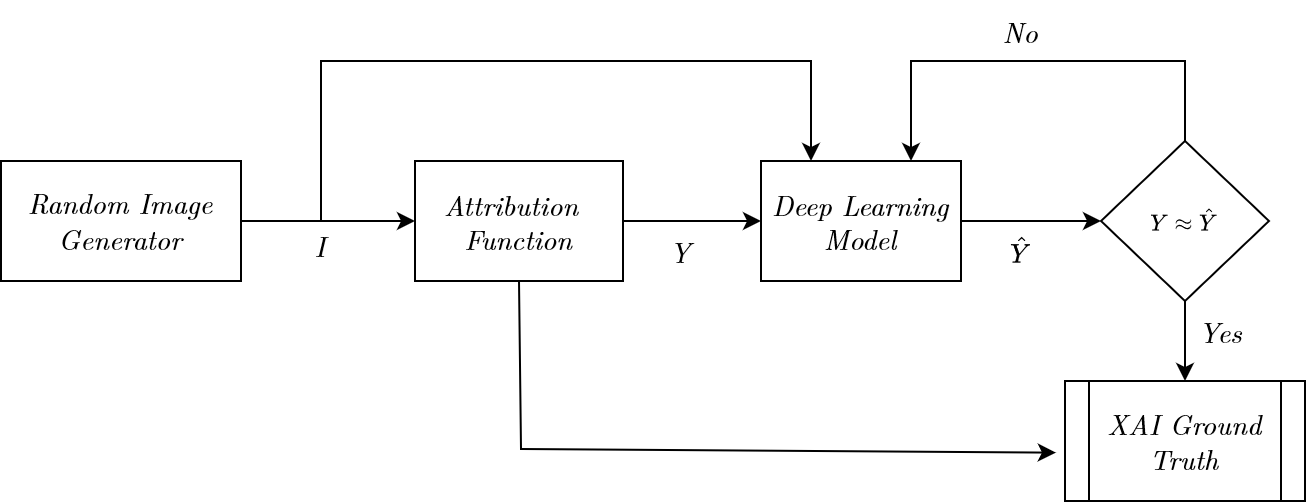}
\end{center}
\caption{Summary of the proposed methodology to generate Synthetic Attribution Benchmark datasets for XAI methods.}\label{fig:aixi_flux}
\end{figure}

All this methodology is summarized in the algorithm \ref{al:method} and figure \ref{fig:aixi_flux}.

In the following section, we describe the experimental setup used to test whether this methodology allowed us to analyse the quality of a given XAI method.

\begin{algorithm}
    \caption{Proposed algorithm to generate synthetic attribution benchmark datasets}\label{al:method}
    \begin{algorithmic}
        \State $I \gets$ Random Image Generator
        \State $Y \gets f(I)$
        \State $\hat{f} \gets $ Initial model
        \State $\hat{Y} \gets  \hat{f}(I)$
        \While{$ Y \not\approx \hat{Y} $}
            \State $\hat{f} \gets$ train($\hat{f}, I, Y$)
            \State $\hat{Y} \gets \hat{f}(I)$
        \EndWhile
    \end{algorithmic}
\end{algorithm}

\section{Experimental setup}
\label{sec:experimental_setup}

The end-goal of the experimental setup was to verify whether the ground truth, generated by the proposed methodology corresponded to the actual explanations. To determine whether this was the case, we conducted multiple experiments. To do these experiments, we needed an XAI method as a proxy to obtain explanations. We need to trust that the results of the XAI method were faithful to the learned relations of the predictive model.

\subsection{Datasets}\label{sec:datasets}

Following the proposed methodology, we generated two different datasets: AIXI-Shape (Artificial Intelligence eXplainable Insights) and AIXI-Color. These datasets consist of 50000 images for training and 2000 for validation. All images had a size of 128 pixels per 128 pixels. We used this image size to simplify the training process. Both the datasets and scripts used to generate them are available at \url{https://github.com/miquelmn/aixi/releases/tag/1.0.0}.

\begin{figure}[!htb]
\centering
\begin{tabular}{cc}
    \subcaptionbox{\label{fig:aixi:a}}{\includegraphics[width=0.4\textwidth]{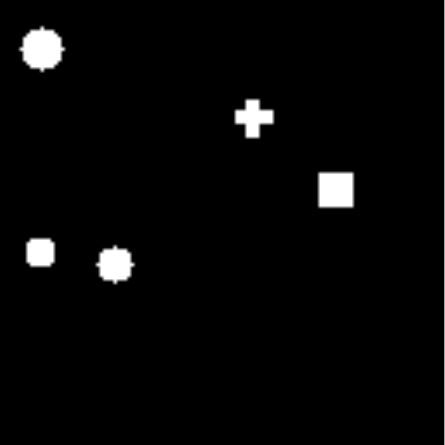}} &  
    \subcaptionbox{\label{fig:aixi:b}}{\includegraphics[width=0.4\textwidth]{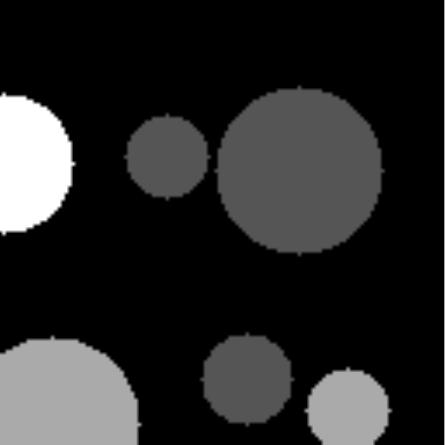}} \\
    \end{tabular}
    \caption{Example of an image from the dataset (a) AIXI-Shape and (b) AIXI-Color.}
    \label{fig:aixi}
\end{figure}

Each dataset allowed us to assess whether the proposed methodology was accurate for patterns of different natures (colours and shapes). 

\subsubsection{AIXI-Shape}
The images from AIXI-Shape were built using a combination of three different patterns: circles, squares, and crosses, as shown in figure \ref{fig:aixi:a}. We used these forms for the different types of visual patterns. We used these simple forms because they are easily recognizable.

These shapes were combined to form each image in the dataset. All the patterns had random sizes, positions, and appearances. Nonetheless, we define a set of restrictions, with the goal to simplify the posterior usage of this dataset:

\begin{itemize}
    \item One image contained a maximum of two objects of the same pattern.
    \item We did not allow overlapping shapes.
    \item All patterns had the same intensity.
\end{itemize}

With the proposed methodology, we must define the function $g(p_{i}, I)$. As already explained in the previous section, this function obtains numerical data from the visual patterns of the images. Here, $g(p_{i}, I) = |p_{i} \in I|$. The position of these, $p_{i}$ in addition to the $w_i$ given by the function used, allows the generation of a GT explanation for each pixel.

\subsubsection{AIXI-Color}

AIXI-Color, in contrast to AIXI-Shape, is a dataset containing only one type of shape, circles, but of different sizes and values. These circles have three possible intensity values, as shown in Figure \ref{fig:aixi:b}. This randomly defined intensity value is the pattern used for the attribution function. Similarly, the positions of the circles were randomized. The random position has only one constraint: the circles must not intersect. This constrain is to avoid possible the existence of completely occluded pattern that could produce miss-prediction and erroneous explanations.

We used the same $g(p_{i}, I)$ of the previous dataset. In addition to the weights of the functions, these features enabled, as already explained for the AIXI-Shape dataset, the generation of ground truth as a saliency map.

\subsection{Functions}\label{sec:functions}

Following the proposed methodology, we must also define the attribution function $F$. All functions, as discussed in the previous section, must comply with a set of requirements. We define three different functions: $ssin$, $suum$ and $class$.

\subsubsection{SSIN}
First proposed by Cortez and Embrechts~\cite{CORTEZ20131} the $ssin$ function (see equation \ref{eq:ssin}), is defined as the sum of three factors based on the $sin$ function. Each of these three factors has an associated weight, $w_i$. These weights were defined by the authors as \textit{theoretical relative importance of} ($0.55$, $0.27$, $0.18$), \textit{e.g., $0.55$ = $0.5/(0.5 + 0.25 + 0.125)$}. This function has one main restriction: the output of the associated function, $g(p_{i}, I)$, must be in the range [0, 1]. This range was defined because the maximum value of $sin$ was obtained using $\pi/2$. Therefore, the maximum value for the $ssin$ function was obtained when all factors had the maximum value of $1$, essentially when $x_1$, $x_2$ and $x_3$ were equal to $\pi/2$.

\begin{equation}
    ssin(x) = 1/2 \cdot sin(\frac{\pi}{2} x_1) + 1/4 \cdot sin(\frac{\pi}{2} x_2) + 1/6 \cdot sin(\frac{\pi}{2} x_3).
    \label{eq:ssin}
\end{equation}

We adapted the original function to our methodology and notation as follows:

\begin{equation}
    ssin(x) = w_{1} \cdot sin(\frac{\pi}{2} g(p_{1}, x)) + w_{2} \cdot sin(\frac{\pi}{2} g(p_{2}, x)) + w_{3} \cdot sin(\frac{\pi}{2} g(p_{3}, x)),
\end{equation}

where $g(p_{i}, x))$ indicates some information of the image $x$ regarding the patten $p_{i}$, and $w_i$ is the weight for each factor, where $w_{1} = 0.55$, $w_{2} = 0.27$, $w_{3} = 0.18$

This function, as stated by Cortez and Embrechts~\cite{CORTEZ20131}, aims to \textit{mimic an additive response of independent nonlinear inputs}. The result was a continuous value between zero and one. For this reason, we considered this to be a regression task.

\subsubsection{SUUM}
We simplify the $ssin$ function to verify whether our methodology can also work with a linear function. We call the resulting function $suum$:

\begin{equation}
    suum(x) = w_1 \cdot g(p_{1}, x) + w_2 \cdot g(p_{2}, x) + w_3 \cdot g(p_{3}, x),
    \label{eq:ssum}
\end{equation}

similar to the $ssin$ function, the weights had the values $w_{1} = 0.55$, $w_{2} = 0.27$, $w_{3} = 0.18$. The theoretical function outputs are in the range of $(-\infty, +\infty)$. We normalized the information output of $g(p_{1}, x)$ to be in the range $[0, 1]$; for this reason, the range of $suum(x)$ was also reduced to $[0, 1]$. Similar to the $ssin$ function, we consider it a regression problem.

\subsubsection{Class}

We define a new function that allows us to simulate a classification problem. To accomplish this, we define a discrete function $class$. This function is defined as follows:

\begin{equation}
    class(x) = w_{1} \cdot g(p_{1}, x) - w_{2} \cdot g(p_{2}, x) \geq 0 ,
    \label{eq:class}
\end{equation}

where $w_1$ and $w_2$ are 1 and 0.5, respectively, $g(p_{i}, x)$ indicates numerical information regarding the visual pattens. We can see, that the function only considered two pattens the rest, if they existed, were ignored. Because the function has a threshold operation, the only possible values are $0$ or $1$. Furthermore, we considered the function of having semantic meaning. For example, if the $g(p_{i}, x)$ function indicates the number of occurrences of pattern $p_{i}$, then the function $class(x)$ is $1$ when there are more objects of the first pattern than of the second. The fact that the output of the function was binary, led us to consider any model that estimates the function as a classification model.

\subsection{Predictive model}\label{sec:predictive}

As explained in previous sections, with this experimental setup, we aimed to verify the quality of the ground truth generated with the proposed methodology. To measure this quality, we must have something to compare it with. For this reason, we required two different elements: a predictive model and an XAI method. We used both for comparison with the proposed GT.

We define a predictive model as a function $f(x)$, where $x$ is a 2D image and $f(x)$ is a prediction of  the content of $x$. Explanations were extracted from the predictive model. We used two types of predictive model. First, we used as the model the functions defined in section \ref{sec:functions}. Second, we used a deep learning model trained to approximate the same functions. 

We first used directly the function as a predictive model to avoid any possible errors in the prediction. We used the explanations obtained from this model as baseline punctuation, which allowed us to identify the limitations of the successive model. Because it was not a real AI model, we could only use agnostic XAI methods. An agnostic method was defined in \cite{ribeiro2016should} as the method that \textit{treat the original model as a black box}.

We also used a deep learning model to simulate a real scenario. Because deep learning models are universal function estimators \cite{hornik1989multilayer}, these types of models can learn any $F$ that we had previously defined. If the model correctly estimated this function, necessarily also had learned the weight associated to each pattern. For this reason, we consider the $w_i$ of each pattern the correct explanation for the result. We used a convolutional neural network (CNN), a special neural network for images-related tasks. The model was reintroduced by Krizhevsky \emph{et al.}~\cite{krizhevsky2012imagenet}. The authors proposed a model with two different parts: a convolutional part, responsible for pattern recognition, and a classification part, responsible for combining these recognized patterns into semantic information. The convolutional part was built with three main elements: convolutional layers, rectified linear unit (ReLU), and max pooling operations. The classification component was a multilayer perceptron (MLP) with a softmax operation as the activation function. All these building blocks were discussed in the original work. These model apply to classification and regression tasks.

\begin{figure}[!htb]
    \centering
    \includegraphics[width=0.8\textwidth]{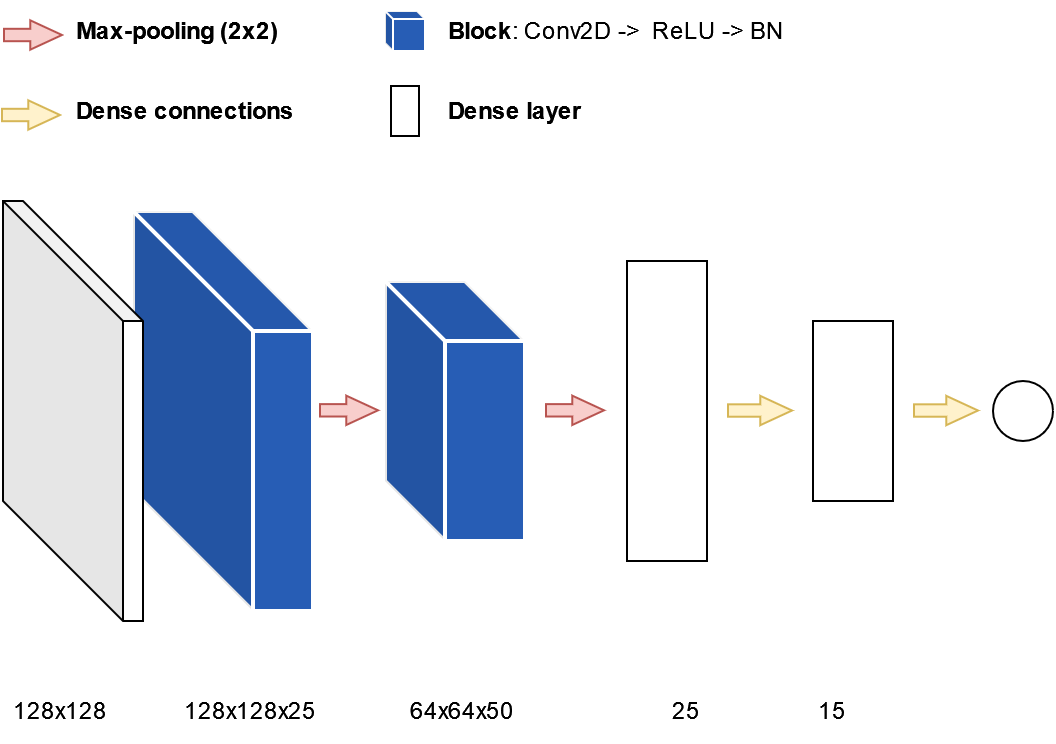}
    \caption{Diagram of the proposed predictive model.}
    \label{fig:aixi_net}
\end{figure}

We developed a compact CNN architecture in response to the straightforward nature of the images within the dataset. We defined this architecture empirically to ensure that the value of the validation metrics was at least on the best decile. We needed to obtain good results to ensure that the AI model learned the desired $F$ function. To construct the feature extractor portion of our model, we utilized two MLP layers. The proposed architecture is illustrated in Figure \ref{fig:aixi_net}.

In both classification and regression tasks, we employed the same architecture, with the only distinction being the addition of a $sigmoid$ activation function for the classification tasks.

In all tasks, we obtained accuracy values for the validation set higher than $0.93$.

\subsection{XAI method}

Once we had defined the predictive models, we also needed to select the XAI method. We used an XAI method to obtain explanations and compare them to our ground truth. Because our goal was to verify the quality of the ground truth and not vice versa, we needed to ensure that the obtained explanations were accurate. We defined a set of requirements to select the best XAI method according to them:

\begin{itemize}
    \item \textbf{Verified}. The XAI method needed to be largely tested and verified to have the minimum number of unknown errors possible. Because it is unlikely to identify a perfect method, the important part of this requirement is to know the possible errors and avoid them.
    \item \textbf{End to end}. The method must be able to analyse the entire AI model. This restriction was set because our proposal generated exact explanations.
    \item \textbf{Fidelity}. Fidelity, as explained in the previous section, is the similarity between the explanation obtained and the actual casual relation learned by the model. Because this experimental setup aimed to verify whether the generated ground truth was correct, we needed to extract the cause of the models with high accuracy to compare it to the ground truth. For this reason, the most important requirement was that the method had a high fidelity.
\end{itemize}

If a method meets all three criteria, we consider that the explanation obtained is correct. Local Interpretable Model-agnostic Explanations (LIME) was the XAI solution that best matched these constraints. 

LIME was first introduced by Ribeiro \emph{et al.}~\cite{ribeiro2016should}. The authors proposed building a local environment around a particular input using occlusions. This local environment of the solution space, allows learning simpler methods, interpretable by design, as a proxy for the underlying, more complex model to be explained.

We verified whether the LIME were complied with the requirements that we defined. First, this method had been widely used, tested, and verified, as demonstrated by over 5000 citations \footnote{Last time checked on 12-01-2022} on \textit{scopus}. Therefore, we considered that a major part of the limitation of the method was already discovered: low performance in the presence of out-of-distribution samples~\cite{qiu2021resisting} and the sensibility to adversarial attacks~\cite{slack_fooling_2020}. Both problems show the lack of robustness of LIME. Robustness, as explained by Alvarez-Melis \emph{et al.}~\cite{alvarez-melis_robustness_2018}, is the expectation that if the data is slightly modified therefore the explanation of this modified data should be similar to the original explanation.  Visani \emph{et al.}~\cite{visani_optilime_2022} identified the origin of this lack of robustness to “the randomness introduced in the sampling step”. All of these problems can be avoided with a well-curated datasets, as the ones we proposed, and small modification to the original LIME implementation. In particular, the fact that at most exist $2^6$ possible samples, due to the maximum number of possible patterns in the images, allowed us to remove the random sampling from the algorithm. Even more, we allowed to select a subset of these possible $2^6$ samples, via a parameter (sample size).

Second, LIME~\cite{ribeiro2016should} is an agnostic method, considering the entire model to obtain its explanations. Finally, as already discussed the sample size defines a trade-off between fidelity and interpretability (or robustness), however and due to the changes we made to the original method, that we already explained, we can increase the overall fidelity of the method without losing robustness. We tested whether this assumption was correct in the experimental setup, analysing how the fidelity increases with a larger sample size. Once we tested it, we maximized the fidelity of the method to ensure that the obtained results provided an exact explanation of the data.

\subsection{Performance measures} \label{sec:metrics}

The results of the explanation method were measured and compared with the ground truth of the datasets. We can find in the literature multiple measures to compare two saliency maps. We based the selection of the measures in the work of Riche \emph{et al.}~\cite{riche2013saliency}, in which the authors review and analyse the state-of-art. Based on their findings, we can deduce that in order to accurately compare two saliency maps, we must treat them as probability distributions. This allowed us to use two well-established measures: 

\begin{itemize}
    \item \textbf{Earth Mover Distance ($\mli{EMD}$)}. First proposed by Rubner \emph{et al.}~\cite{rubner1998metric}. $\mli{EMD}$ can be defined as the minimum amount of work required to transform one distribution into the other. See equation \ref{eq:emd}.

    \begin{equation}
    \begin{aligned}
        \mli{EMD} = {} & (\min_{f_{ij}} \sum_{i,j} f_{ij} \cdot d_{ij}) + | \sum_{i} \mli{SM}_{i} - \sum_{j} \hat{\mli{SM}_{j}}| \max_{i,j} d_{i,j} \\
                & s.t. f_{i,j} \geq 0, \sum_{j} f_{i,j} \leq \mli{SM}_{i}, \sum_{i} f_{i,j} \leq \hat{\mli{SM}_{j}} \\
                & \sum_{i, j} f_{i, j} = \min(\sum_{i} \mli{SM}_{i} - \sum_{j} \hat{\mli{SM}_{j}})
    \end{aligned}
    \label{eq:emd}
    \end{equation}
    
    \noindent where each $f_{ij}$ represents the amount transported from the $i_{th}$ supply to the $j_{th}$ demand. $d_{ij}$ is the ground distance between the sample $i$ and sample $j$ in the distribution. $\mli{SM}$ is the probability distribution of the ground truths and $\hat{\mli{SM}}$ the probability distribution of the predicted saliency map. 
    
    \item \textbf{Kullback–Leibler divergence ($\mli{KL}_{div}$)}. First proposed by Kullback and Leibler~\cite{kullback1951information}. The $\mli{KL}_{div}$ is a measure of the information lost from one distribution to the other. See equation \ref{eq:kl}.

    \begin{equation}
        \begin{aligned}
            \mli{KL}_{div} = \sum^{X}_{x=1} \mli{SM}(x) \cdot \mli{log}(\frac{\mli{SM(x)}}{\hat{\mli{SM(x)} + \epsilon}} + \epsilon),
        \end{aligned}
        \label{eq:kl}
    \end{equation}

    \noindent where $\mli{SM}$ is the probability distribution of the ground truths and $\hat{\mli{SM}}$ the probability distribution of the predicted saliency map. 

\end{itemize}

Both measures have the best result equal to $0$, however while $\mli{EMD}$ have a lower boundary with value $1$ as the worst possible results, $\mli{KL}_{div}$ do not have any boundary. This fact is important for any further analysis.

\subsection{Experiments} \label{sec:experiments}

Three experiments were conducted to study the proposed methodology. The first experiment tests a basic assumption of the rest of experiments, that the fidelity of LIME~\cite{ribeiro2016should} can be increased through the size of the sample. In the second experiment, we compared the proposed ground truth with the results of LIME~\cite{ribeiro2016should}, configured to obtain the maximum fidelity possible. In this experiment, we used the function itself instead of an AI model. Finally, in the third experiment, we did the same as in the second experiment but with an AI model. The last two experiments had six sub-experiments each, all of these sub-experiments share that they were built with a combination of one dataset and a function.

These experiments aimed to assess whether the GT generated by our methodology was in fact the correct ground truth. 

In all experiments, we used LIME as the XAI method, to obtain explanations. We compared these explanations with the GT generated by our method. We used both metrics defined in Section \ref{sec:metrics}. However, as already discussed, the $\mli{KL}_{div}$ measure does not have a reference value to be able to further analyse the results. We used as a baseline value the results obtained with the smallest possible sample size calculated in the first experiment. 

In the second experiment, we used a function $F$ as the predictor to avoid unexpected errors provoked by the AI model. Because we did not train any model, we could be sure that the prediction was perfect, and for this reason, we know that all errors were produced either by the XAI method or the proposed methodology. We considered the results obtained from these experiments as baseline results. However, because the $class(x)$ function is a binary and the result value only varies if there is a big enough perturbation, we expect that the explanations from the AI model surpass in fidelity this function. This counter-intuitive behaviour can be even exaggerated if the sigmoid activation function of the last layer is removed.

In the third experiment, we used an AI model. We use the model explained in Section \ref{sec:predictive}. This experiment aimed to simulate a real scenario in a simplified manner. Similar to the second experiment, we extracted the explanations using LIME~\cite{ribeiro2016should}. The major caveat of using an AI model was the possibility of adding new errors without relation to our methodology.

These two last experiments were divided into six sub-experiment. Each of these six sub-experiments was defined by the dataset and function used, as presented in sections \ref{sec:datasets} and \ref{sec:functions}:

\begin{itemize}
    \item \textbf{Sub-experiments $\bm{ssin}$}. These experiments aimed to test the proposed methodology for the regression of non-linear functions. We defined two sub-experiments with different datasets that aimed to test this. We used, in both cases, the $ssin$ function. 
    
    We used in the first sub-experiment the \textbf{AIXI-Shape} dataset and in the second the \textbf{AIXI-Color}. In both cases, the $g(p_{i}, I)$ used to calculate the $ssin$ was $g(p_{i}, I) = |p_{i} \in I|$, indicating how many objects of the respective pattern were found on the image.
    
    \item \textbf{Sub-experiments $\bm{suum}$}. These experiments were defined to assess whether our methodology was useful for the regression of linear functions. We used the same two datasets from the previous experiments. We used the $suum$ function and again $g(p_{i}, I) = |p_{i} \in I|$.

   \item \textbf{Sub-experiments $\bm{class}$}. These experiments were defined to assess whether our methodology was useful for classification tasks. We used the $class$ function, defined in Section \ref{sec:functions}, and similarly to the previous experiments, both \textbf{AIXI-Shape} and \textbf{AIXI-Color} dataset and $g(p_{i}, I) = |p_{i} \in I|$. 
\end{itemize}

\section{Results and discussion}\label{sec:results}

In this section, we discuss and analyse the results of the experiments defined in Section \ref{sec:experiments}. 

\subsection{Experiment 1}

\begin{figure}[h!]
	\centering
    	\subfloat[Value of $\mli{EMD}$ between the LIME~\cite{ribeiro2016should} results and the different datasets. The horizontal axis indicated the size of the sampling process.]{\includegraphics[width=0.7\textwidth]{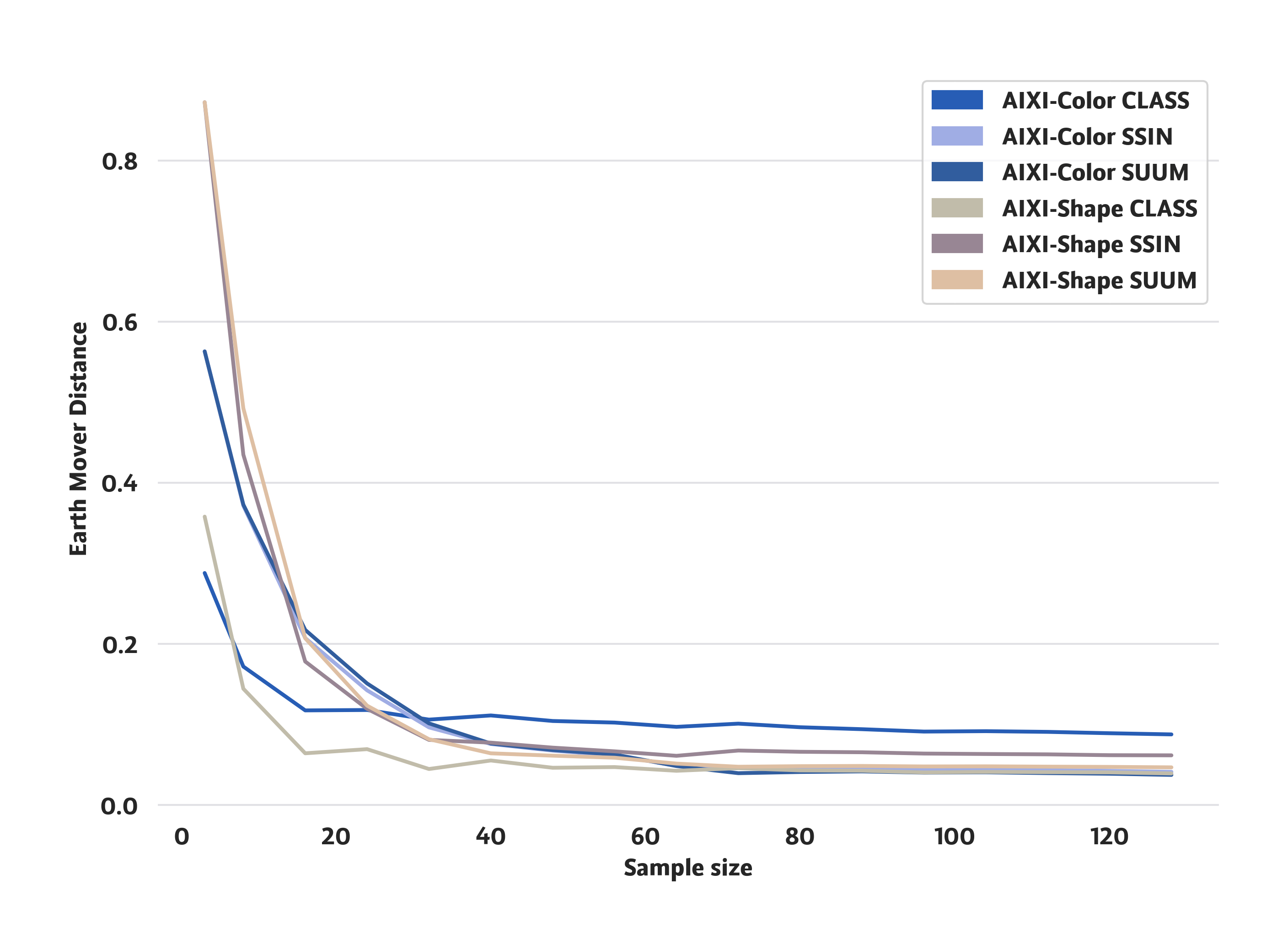}%
    	\label{subfigure:ex_1_emd}}
    	\hfil
    	\subfloat[Value of $\mli{KL}_{div}$ between the LIME~\cite{ribeiro2016should} results and the different datasets. The horizontal axis indicated the size of the sampling process.]{\includegraphics[width=0.7\textwidth]{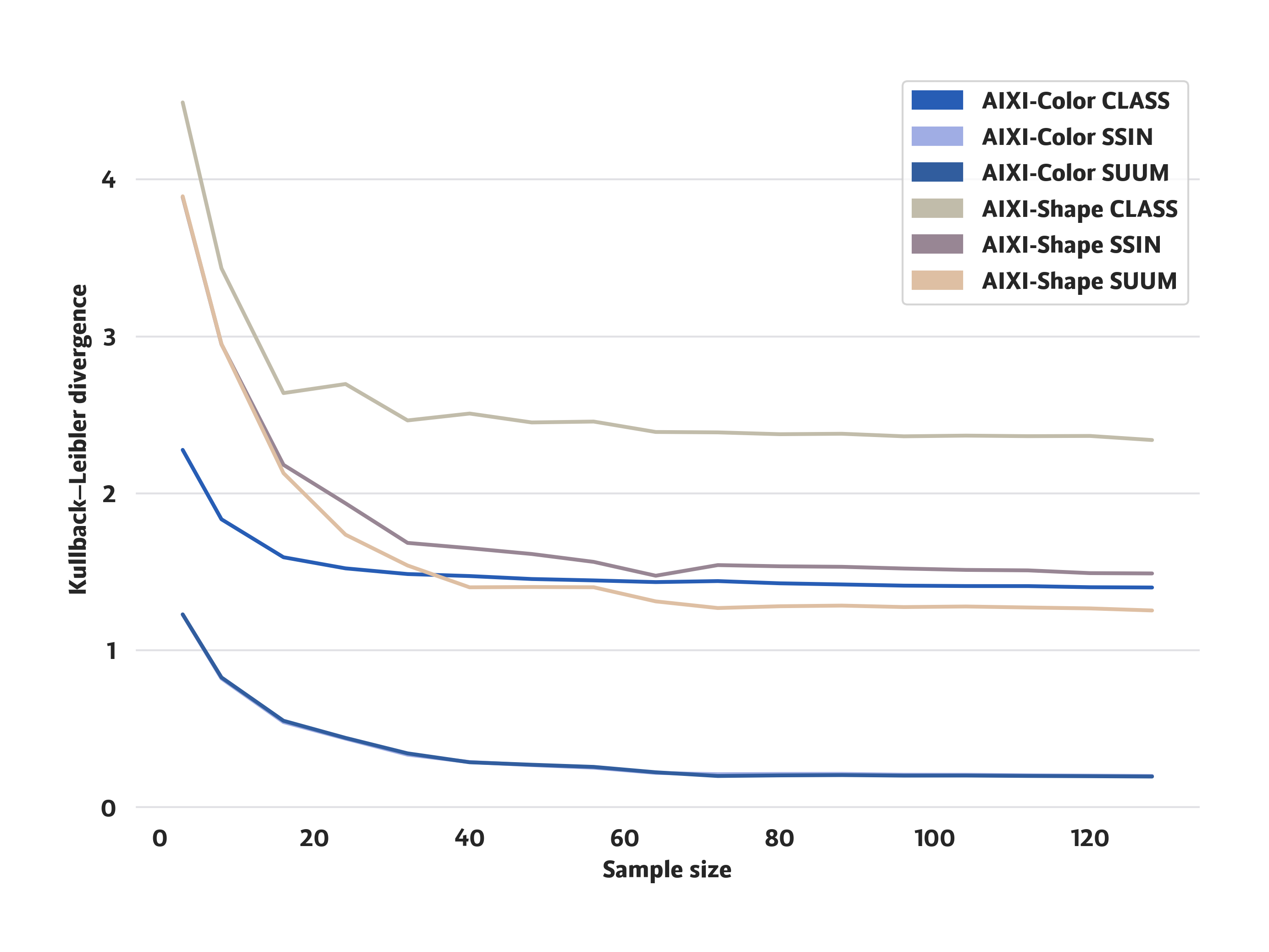}%
    	\label{subfigure:ex_1_kl}}
     \caption{Results of the first experiment.}
	\label{fig:neighbour}
\end{figure}

Figure \ref{fig:neighbour} depicts the $\mli{EMD}$ and $\mli{KL}_{div}$ results to compare our proposed ground truth and the saliency map results from LIME~\cite{ribeiro2016should} with different sampling size. 

We can observe, for both metrics, that, clearly, with an increased sample size we obtain results more similar to our ground truth. These results are consistent with the original work from Ribeiro \emph{et al.}~\cite{ribeiro2016should} that expects an increase in the size of the sample generates explanations with higher fidelity. In addition, and because we removed the stochasticity of the sampling process, as already explained in the previous section, the danger of a lack of robustness with an increased sampling is removed. We realized all the existing perturbations, $2^6$, and we obtained its corresponding explanations.

The results of this experiment confirm that with our modified LIME~\cite{ribeiro2016should} proposal, we can obtain high fidelity values while maintaining the robustness of the method. These two facts allowed us to realize further experimentation, knowing that the results of LIME~\cite{ribeiro2016should} (when the sampling process uses all possible samples) are a good comparison element to know whether our ground truth is correct or not. In addition, we used the results of LIME with the minimum samples obtained in this experiment as the baseline results to compare the results of the rest of experiments. This baseline values can be seen in table \ref{tab:baseline}.

\begin{table}[!htb]
\centering
\begin{tabular}{llcc}
\hline
Function                    & Dataset           & $\mli{KL}_{div}$  &  $\mli{EMD}$   \\
\hline
\multirow{2}{*}{$SSIN$}     & AIXI-Shape        & 3.8873            & 0.8722        \\     
                            & AIXI-Color        & 1.2284            & 0.5631        \\ \hline 
\multirow{2}{*}{$SUUM$}     & AIXI-Shape        & 3.8926            & 0.8722        \\
                            & AIXI-Color        & 1.2286            & 0.5631        \\ \hline
\multirow{2}{*}{$CLASS$}    & AIXI-Shape        & 4.4907            & 0.3579        \\
                            & AIXI-Color        & 2.2768            & 0.2880        \\
\hline
\end{tabular}
\caption{Results obtained for experiment 1 with the minimum sample size. These values can be used as a baseline method.}
\label{tab:baseline}
\end{table}

\subsection{Experiment 2}

Table \ref{tab:results_second_experiment} shows the $\mli{KL}_{div}$ and $\mli{EMD}$ results obtained within the second experiment. 

We observe that the $\mli{EMD}$, are nearly approaching 0, indicating a high similarity between our ground truth and the outcomes generated by LIME~\cite{ribeiro2016should}. Similarly, we can see that $\mli{KL}_{div}$ have low values.  Evaluating this latter metric is somewhat intricate due to the requirement for a baseline value. We used the results obtained in the first experiment as the baseline (see table \ref{tab:baseline}), and we can see that from the LIME~\cite{ribeiro2016should} with less fidelity to the one used in this experiment the distance is reduced by two points. 

We can also see, in both metrics results, that the worst results obtained were with the $class$ function. As already explained, we expected this behaviour to the limitation of the occluding process in a scenario where it can exist perturbation that do not modify in any degree the output value. We analysed the non-agreggated results, and we detected that the images with worse results were the ones that even occluding all the objects did not provoke change on the function output. Nonetheless, even the results of the classification tasks had nearly perfect measures.  

\begin{table}[!htb]
\centering
\begin{tabular}{llcc}
\hline
Function                    & Dataset           & $\mli{KL}_{div}$  &  $\mli{EMD}$   \\
\hline
\multirow{2}{*}{$SSIN$}     & AIXI-Shape        & 1.4895            & 0.0618    \\     
                            & AIXI-Color        & 0.1993            & 0.0414    \\ \hline 
\multirow{2}{*}{$SUUM$}     & AIXI-Shape        & 1.2537            & 0.0469    \\
                            & AIXI-Color        & 0.1954            & 0.0375    \\ \hline
\multirow{2}{*}{$CLASS$}    & AIXI-Shape        & 2.3396            & 0.0394    \\
                            & AIXI-Color        & 1.4000            & 0.0878    \\
\hline
\end{tabular}
\caption{Results obtained for experiment 2. The values indicated the result obtained with a sample size of 128.}
\label{tab:results_second_experiment}
\end{table}

Based on the previous analysis, we can state that the proposed methodology generates the ground truth very similar to the explanations obtained from a high fidelity and highly robust LIME~\cite{ribeiro2016should}. Even so, there were sub-experiments with worse results, we expected that behaviour, and we assessed that it was provoked by the limitation of the method used. For this reason, we conclude that, in this controlled environment we generated a correct ground truth for the explanations. In the following experiment, we analyse whether these results are consistent in a more realistic scenario. 

\subsection{Experiment 3}

Table \ref{tab:results_second_experiment} lists the results obtained in Experiment 3. Similar to the previous experiment, this table summarizes the $\mli{KL}_{div}$ and $\mli{EMD}$ results from all sub-experiments. This experiment used an AI model as the predictive model that estimates the respective $F: \mathcal{X} \to \mathbb{R}$ function.

We can see that the results, in the regression tasks, are worse than the previous experiment. We expect this behaviour to the possible inclusion of errors from the neural network. However, we can see that, in the classification tasks, the results are slightly better than the previous experiment. We already explained that we also expected this behaviour to the sensitivity of this method to small changes, in comparison to the function inability to be affected by small changes.

\begin{table}[!htb]
\centering
\begin{tabular}{llccccl}
\hline
Function                    & Dataset           & $\mli{KL}_{div}$  &  $\mli{EMD}$  \\
\hline
\multirow{2}{*}{$SSIN$}     & AIXI-Shape        & 1.6722            & 0.0728        \\ 
                            & AIXI-Color        & 0.3029            & 0.0784        \\ 
                            \hline
\multirow{2}{*}{$SUUM$}     & AIXI-Shape        & 1.6708            & 0.0734        \\
                            & AIXI-Color        & 0.3175            & 0.0854        \\
                            \hline
\multirow{2}{*}{$CLASS$}    & AIXI-Shape        & 1.5751            & 0.0679        \\ 
                            & AIXI-Color        & 1.2971            & 0.0856        \\
\hline
\end{tabular}
\caption{Results obtained for experiment 3. The values indicated the result obtained with a sample size of 128.}
\label{tab:results_second_experiment}
\end{table}

Although the results have a small diverge from the previous experiment, we can see that are still almost perfect results. This fact is more easily concluded from the $\mli{EMD}$ results. All sub-experiments had results near to 0. Furthermore, the $\mli{KL}_{div}$ had also small values in comparison to the baseline from the first experiment. 

Based on the previous analysis, we can state that our methodology is consistent in a more realistic environment. The fact that the results of this third experiment were almost perfect, and for this reason also similar to those of the second experiment, indicated that the AI model approximates correctly the function $F$. The above demonstrates that our assumptions hold: when we used the proposed methodology, the GT generated was the actual cause for the predictions for the AI models.

\color{black}
\section{Conclusion}
\label{sec:conclusion}

In this study, we propose a novel methodology to generate a Synthetic attribution benchmark (SAB) dataset of images. Our methodology allowed us to define the precise importance for each pixel of an image and can be used to verify most of XAI techniques, methods, and metrics equally, proposed in the literature. This ground truth (GT) allowed us to ignore the ad hoc solutions proposed in the state-of-art. 

We verified through three different experiments, both in a controlled environment, the correctness of this approach. We obtained explanations using LIME~\cite{ribeiro2016should} to verify whether our GT was correct. We used this algorithm because its results achieved high fidelity, as demonstrated in the first experiment. Our GT was very similar to the LIME~\cite{ribeiro2016should} results, when this method was forced to be both robust and faithful to the real cause of the prediction. Therefore, we can state that we generated the correct ground truth.

We included as results the implemented code for the generation of the datasets (see \url{https://github.com/miquelmn/aixi/releases/tag/1.0.0}). As an additional contribution we constructed and opened to the scientific community two synthetic datasets, AIXI-Color and AIXI-Value, with the positions of the objects and their classes as a ground truth. For the sake of scientific progress, it would be beneficial if authors published their raw data, code, and the image datasets.

\section{Declaration of competing interest}

The authors declare that they have no known competing financial interests or personal relationships that could have influenced the work reported in this study.



\section{Funding}
Project PID2019-104829RA-I00 “EXPLainable Artificial INtelligence systems for health and well-beING (EXPLAINING)” funded by \\ MCIN/AEI/10.13039/501100011033. Miquel Miró-Nicolau benefited from the fellowship FPI\_035\_2020 from Govern de les Illes Balears.

\bibliographystyle{abbrvnat} 
\bibliography{references}

\begin{thebibliography}{22}
\providecommand{\natexlab}[1]{#1}
\providecommand{\url}[1]{\texttt{#1}}
\expandafter\ifx\csname urlstyle\endcsname\relax
  \providecommand{\doi}[1]{doi: #1}\else
  \providecommand{\doi}{doi: \begingroup \urlstyle{rm}\Url}\fi

\bibitem[Adebayo et~al.()Adebayo, Gilmer, Muelly, Goodfellow, Hardt, and
  Kim]{adebayo_sanity_2018}
J.~Adebayo, J.~Gilmer, M.~Muelly, I.~Goodfellow, M.~Hardt, and B.~Kim.
\newblock Sanity checks for saliency maps.
\newblock 2018-Decem:\penalty0 9505--9515.
\newblock ISSN 10495258.

\bibitem[Alvarez-Melis and Jaakkola()]{alvarez-melis_robustness_2018}
D.~Alvarez-Melis and T.~S. Jaakkola.
\newblock On the robustness of interpretability methods.
\newblock URL \url{http://arxiv.org/abs/1806.08049}.

\bibitem[Arras et~al.(2022)Arras, Osman, and Samek]{arras2020ground}
L.~Arras, A.~Osman, and W.~Samek.
\newblock Clevr-xai: A benchmark dataset for the ground truth evaluation of
  neural network explanations.
\newblock \emph{Information Fusion}, 81:\penalty0 14--40, 2022.
\newblock ISSN 1566-2535.
\newblock \doi{https://doi.org/10.1016/j.inffus.2021.11.008}.
\newblock URL
  \url{https://www.sciencedirect.com/science/article/pii/S1566253521002335}.

\bibitem[Barredo~Arrieta et~al.()Barredo~Arrieta, Díaz-Rodríguez, Del~Ser,
  Bennetot, Tabik, Barbado, Garcia, Gil-Lopez, Molina, Benjamins, Chatila, and
  Herrera]{barredo_arrieta_explainable_2020}
A.~Barredo~Arrieta, N.~Díaz-Rodríguez, J.~Del~Ser, A.~Bennetot, S.~Tabik,
  A.~Barbado, S.~Garcia, S.~Gil-Lopez, D.~Molina, R.~Benjamins, R.~Chatila, and
  F.~Herrera.
\newblock Explainable artificial intelligence ({XAI}): Concepts, taxonomies,
  opportunities and challenges toward responsible {AI}.
\newblock 58:\penalty0 82--115.
\newblock ISSN 15662535.
\newblock \doi{10.1016/j.inffus.2019.12.012}.
\newblock URL
  \url{https://linkinghub.elsevier.com/retrieve/pii/S1566253519308103}.

\bibitem[Cortez and Embrechts(2013)]{CORTEZ20131}
P.~Cortez and M.~J. Embrechts.
\newblock Using sensitivity analysis and visualization techniques to open black
  box data mining models.
\newblock \emph{Information Sciences}, 225:\penalty0 1--17, 2013.
\newblock ISSN 0020-0255.
\newblock \doi{https://doi.org/10.1016/j.ins.2012.10.039}.
\newblock URL
  \url{https://www.sciencedirect.com/science/article/pii/S0020025512007098}.

\bibitem[Guidotti(2021)]{guidotti_evaluating_2021}
R.~Guidotti.
\newblock Evaluating local explanation methods on ground truth.
\newblock \emph{Artificial Intelligence}, 291:\penalty0 103428, Feb. 2021.
\newblock ISSN 00043702.
\newblock \doi{10.1016/j.artint.2020.103428}.
\newblock URL
  \url{https://linkinghub.elsevier.com/retrieve/pii/S0004370220301776}.

\bibitem[Hornik et~al.(1989)Hornik, Stinchcombe, and
  White]{hornik1989multilayer}
K.~Hornik, M.~Stinchcombe, and H.~White.
\newblock Multilayer feedforward networks are universal approximators.
\newblock \emph{Neural networks}, 2\penalty0 (5):\penalty0 359--366, 1989.

\bibitem[Krizhevsky et~al.(2012)Krizhevsky, Sutskever, and
  Hinton]{krizhevsky2012imagenet}
A.~Krizhevsky, I.~Sutskever, and G.~E. Hinton.
\newblock Imagenet classification with deep convolutional neural networks.
\newblock \emph{Advances in neural information processing systems}, 25, 2012.

\bibitem[Kullback and Leibler(1951)]{kullback1951information}
S.~Kullback and R.~A. Leibler.
\newblock On information and sufficiency.
\newblock \emph{The annals of mathematical statistics}, 22\penalty0
  (1):\penalty0 79--86, 1951.

\bibitem[Mamalakis et~al.(2022)Mamalakis, Barnes, and
  Ebert-Uphoff]{mamalakis2022investigating}
A.~Mamalakis, E.~A. Barnes, and I.~Ebert-Uphoff.
\newblock Investigating the fidelity of explainable artificial intelligence
  methods for applications of convolutional neural networks in geoscience.
\newblock \emph{Artificial Intelligence for the Earth Systems}, 1\penalty0
  (4):\penalty0 e220012, 2022.

\bibitem[Miller(2019)]{miller2019explanation}
T.~Miller.
\newblock Explanation in artificial intelligence: Insights from the social
  sciences.
\newblock \emph{Artificial intelligence}, 267:\penalty0 1--38, 2019.

\bibitem[Mir{\'o}-Nicolau et~al.(2022)Mir{\'o}-Nicolau, Moy{\`a}-Alcover, and
  Jaume-i Cap{\'o}]{miro2022evaluating}
M.~Mir{\'o}-Nicolau, G.~Moy{\`a}-Alcover, and A.~Jaume-i Cap{\'o}.
\newblock Evaluating explainable artificial intelligence for x-ray image
  analysis.
\newblock \emph{Applied Sciences}, 12\penalty0 (9):\penalty0 4459, 2022.

\bibitem[Murdoch et~al.()Murdoch, Singh, Kumbier, Abbasi-Asl, and
  Yu]{murdoch_definitions_2019}
W.~J. Murdoch, C.~Singh, K.~Kumbier, R.~Abbasi-Asl, and B.~Yu.
\newblock Definitions, methods, and applications in interpretable machine
  learning.
\newblock 116\penalty0 (44):\penalty0 22071--22080.
\newblock ISSN 10916490.
\newblock \doi{10.1073/pnas.1900654116}.

\bibitem[Qiu et~al.(2022)Qiu, Yang, Cao, Zheng, Ngai, Hsiao, and
  Chen]{qiu2021resisting}
L.~Qiu, Y.~Yang, C.~C. Cao, Y.~Zheng, H.~Ngai, J.~Hsiao, and L.~Chen.
\newblock Generating perturbation-based explanations with robustness to
  out-of-distribution data.
\newblock In \emph{Proceedings of the ACM Web Conference 2022}, pages
  3594--3605, 2022.

\bibitem[Ribeiro et~al.(2016)Ribeiro, Singh, and Guestrin]{ribeiro2016should}
M.~T. Ribeiro, S.~Singh, and C.~Guestrin.
\newblock " why should i trust you?" explaining the predictions of any
  classifier.
\newblock In \emph{Proceedings of the 22nd ACM SIGKDD international conference
  on knowledge discovery and data mining}, pages 1135--1144, 2016.

\bibitem[Riche et~al.(2013)Riche, Duvinage, Mancas, Gosselin, and
  Dutoit]{riche2013saliency}
N.~Riche, M.~Duvinage, M.~Mancas, B.~Gosselin, and T.~Dutoit.
\newblock Saliency and human fixations: State-of-the-art and study of
  comparison metrics.
\newblock In \emph{Proceedings of the IEEE international conference on computer
  vision}, pages 1153--1160, 2013.

\bibitem[Rubner et~al.(1998)Rubner, Tomasi, and Guibas]{rubner1998metric}
Y.~Rubner, C.~Tomasi, and L.~J. Guibas.
\newblock A metric for distributions with applications to image databases.
\newblock In \emph{Sixth international conference on computer vision (IEEE Cat.
  No. 98CH36271)}, pages 59--66. IEEE, 1998.

\bibitem[Slack et~al.()Slack, Hilgard, Jia, Singh, and
  Lakkaraju]{slack_fooling_2020}
D.~Slack, S.~Hilgard, E.~Jia, S.~Singh, and H.~Lakkaraju.
\newblock Fooling {LIME} and {SHAP}: Adversarial attacks on post hoc
  explanation methods.
\newblock In \emph{Proceedings of the {AAAI}/{ACM} Conference on {AI}, Ethics,
  and Society}, pages 180--186. {ACM}.
\newblock ISBN 978-1-4503-7110-0.
\newblock \doi{10.1145/3375627.3375830}.
\newblock URL \url{https://dl.acm.org/doi/10.1145/3375627.3375830}.

\bibitem[Tomsett et~al.(2020)Tomsett, Harborne, Chakraborty, Gurram, and
  Preece]{tomsett2020sanity}
R.~Tomsett, D.~Harborne, S.~Chakraborty, P.~Gurram, and A.~Preece.
\newblock Sanity checks for saliency metrics.
\newblock In \emph{Proceedings of the AAAI conference on artificial
  intelligence}, volume~34, pages 6021--6029, 2020.

\bibitem[van~der Velden et~al.(2022)van~der Velden, Kuijf, Gilhuijs, and
  Viergever]{VanderVelden2021}
B.~H. van~der Velden, H.~J. Kuijf, K.~G. Gilhuijs, and M.~A. Viergever.
\newblock {Explainable artificial intelligence (XAI) in deep learning-based
  medical image analysis}.
\newblock \emph{Medical Image Analysis}, 79:\penalty0 102470, jul 2022.
\newblock ISSN 13618415.
\newblock \doi{10.1016/j.media.2022.102470}.
\newblock URL \url{http://arxiv.org/abs/2107.10912
  https://linkinghub.elsevier.com/retrieve/pii/S1361841522001177}.

\bibitem[Vaswani et~al.(2017)Vaswani, Shazeer, Parmar, Uszkoreit, Jones, Gomez,
  Kaiser, and Polosukhin]{vaswani2017attention}
A.~Vaswani, N.~Shazeer, N.~Parmar, J.~Uszkoreit, L.~Jones, A.~N. Gomez,
  {\L}.~Kaiser, and I.~Polosukhin.
\newblock Attention is all you need.
\newblock \emph{Advances in neural information processing systems}, 30, 2017.

\bibitem[Visani et~al.()Visani, Bagli, and Chesani]{visani_optilime_2022}
G.~Visani, E.~Bagli, and F.~Chesani.
\newblock {OptiLIME}: Optimized {LIME} explanations for diagnostic computer
  algorithms.
\newblock URL \url{http://arxiv.org/abs/2006.05714}.

\end{thebibliography}





\end{document}